\documentclass[letterpaper, 10 pt, journal, twoside]{ieeetran}  

\usepackage{amssymb,amsmath,graphicx,color,algorithm,algorithmic,url}
\usepackage[english]{babel}
\usepackage[noadjust]{cite} 
\usepackage{hhline,caption}
\DeclareMathOperator*{\mm}{{\rm mm}}

\IEEEoverridecommandlockouts                              

\title{
From pixels to percepts: Highly robust edge perception and contour following using deep learning and an optical biomimetic tactile sensor 
\vspace{0em}
}

\author{Nathan F. Lepora$^*$, Alex Church, Conrad De Kerckhove, Raia Hadsell, John Lloyd$^*$%
\thanks{Manuscript received: September 10, 2018; Revised December 6, 2018; Accepted January 23, 2019.}
\thanks{This paper was recommended for publication by Editor Dan Popa upon evaluation of the Associate Editor and Reviewers' comments. This work was supported by an award from the Leverhulme Trust on ‘A biomimetic forebrain for robot touch’ (RL-2016-39). * NL and JL contributed equally to this work.} 
\thanks{$^{1}$NL, AC, CDK and JL are with the Department of Engineering Mathematics and Bristol Robotics Laboratory, University of Bristol, Bristol, U.K.
	{\tt\footnotesize \{n.lepora, ac14293, cd14192, jl15313\}@bristol.ac.uk}}%
\thanks{$^{2} $RH is with Google DeepMind. 
	{\tt\footnotesize raia@google.com}}%
\thanks{Digital Object Identifier (DOI): see top of this page.}
}

\begin{document}

\maketitle

\begin{abstract}
Deep learning has the potential to have the impact on robot touch that it has had on robot vision. Optical tactile sensors act as a bridge between the subjects by allowing techniques from vision to be applied to touch. In this paper, we apply deep learning to an optical biomimetic tactile sensor, the TacTip, which images an array of papillae (pins) inside its sensing surface analogous to structures within human skin. {\color{black}Our main result is that the application of a deep CNN can give reliable edge perception and thus a robust policy for planning contact points to move around object contours.} Robustness is demonstrated over several irregular and compliant objects with both tapping and continuous sliding, using a model trained only by tapping onto a disk. These results relied on using techniques to encourage generalization to tasks beyond which the model was trained. We expect this is a generic problem in practical applications of tactile sensing that deep learning will solve.
\end{abstract}


\section{INTRODUCTION}

\IEEEPARstart{R}{obot} touch differs from robot vision: to touch, an agent must physically interact with its environment, which constrains the form and function of its tactile sensors (for example, to be robust, compliant and compact). Likewise, tactile perception differs from visual perception: to perceive touch, an agent interprets the deformation of its sensing surface, which depends on the shape and mechanics of the sensor, unlike vison where the eye does change what can potentially be seen~\cite{Hayward2011}. 

Therefore, the application of deep learning to robot touch will be different from robot vision, just as robot vision poses different research questions than computer vision~\cite{Sunderhauf2018a}. Thus far, there have been relatively few studies of deep learning for tactile perception compared with the explosion of work on robot vision. Those studies have mainly considered one particular device -- the Gelsight~\cite{Yuan2017}, an optical tactile sensor that images the shading from 3 internal RGB-colored LEDs to transduce surface deformation (and complements this with painted markers to detect shear~\cite{Yuan2015}). Use of an optical tactile sensor seems an appropriate starting place for applying deep learning to touch, given the rapid progress for vision. 

\begin{figure}[t]
	\centering
	\begin{tabular}[b]{@{}cc@{}}
		{\bf (a) Robot: Arm-mounted sensor} & {\bf (b) Tactile sensor} \\
		\includegraphics[width=0.6\columnwidth,trim={0 0 0 0},clip]{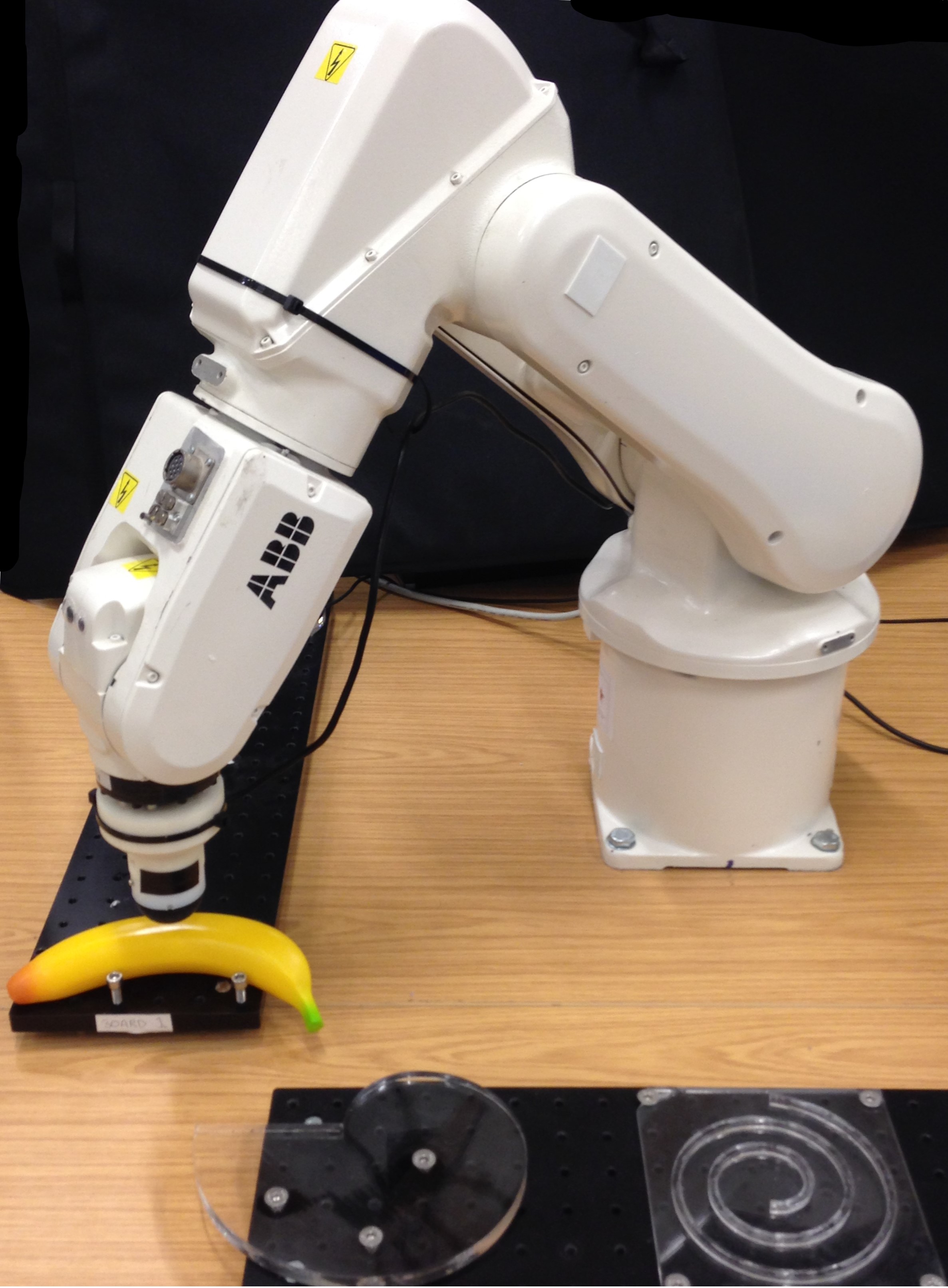} & 
		\begin{tabular}[b]{@{}c@{}}
			\includegraphics[width=0.30\columnwidth,trim={0 10 0 25},clip]{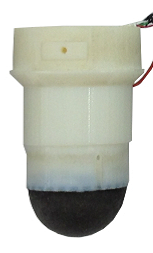} \\
			{\bf (c) Tactile image} \\
			\includegraphics[width=0.35\columnwidth,trim={15 10 0 0},clip]{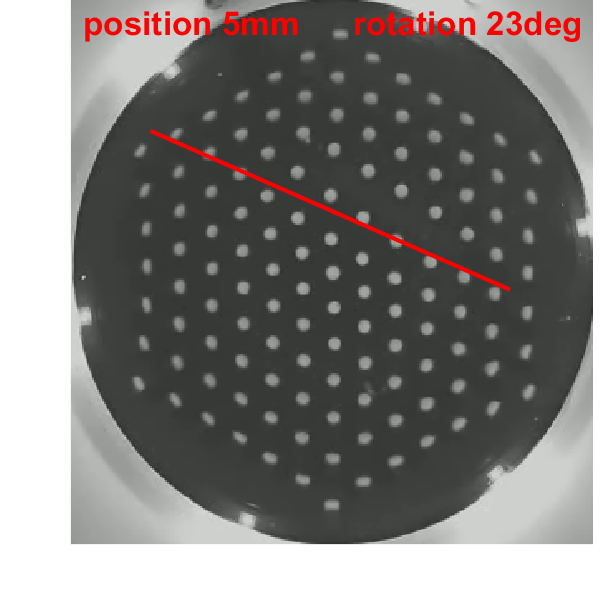}
		\end{tabular}
	\end{tabular}
	\vspace{-1em}
	\caption{Tactile robotic system, showing: (a) the arm-mounted tactile sensor next to a few objects; (b) the (TacTip) optical tactile sensor; and (c) a labelled image from the tactile sensor.}
	\label{fig:1}
	\vspace{-1em}
\end{figure} 

In this paper, we apply deep learning to an optical biomimetic tactile sensor, the TacTip~\cite{Chorley2009,Ward-Cherrier2018}, which images an array of 3D pins inside its sensing surface (Fig.~\ref{fig:1}c). {\color{black}These pins mimic the papillae that protrude across the dermal/epidermal layers of tactile skin, within which mechanoreceptors sense their displacement~\cite{Cramphorn2017}.} In past work, we used image-processing to track the pin positions~\cite{Lepora2015,Lepora2016}. These were then passed to a distinct perception system, using statistical models~\cite{Lepora2015,Lepora2016,Lepora2017,Ward-Cherrier2018,Ward-Cherrier2017a,Cramphorn2017,Cramphorn2016}, dimensionality reduction~\cite{Aquilina2018} or support vector machines~\cite{James2018}. Here we use a deep Convolutional Neural Network (CNN) for direct (end-to-end) perception from the tactile images. 

\textcolor{black}{Our main result is that application of a deep CNN can give robust edge perception and contour following in tasks beyond which the model was trained. We define a policy for planning contacts that moves the sensor while maintaining its pose relative to the edge. We test the robustness of this policy by whether the robot can successfully complete closed contours: first, with a tapping motion around a disk under variations in contact trajectory; second, around irregular objects differing in local shape (Fig.~\ref{fig:1}); and third, for continual sliding motion. In all cases, the model was trained by tapping on a region of the disk. The policy met these tests, struggling only when sliding around sharp corners.}


\section{BACKGROUND AND RELATED WORK}

Initial applications of deep learning to artificial tactile perception were with taxel-based sensors. The first used a four-digit robot hand covered with 241 distributed tactile skin sensors (with another 71 motor angles, currents and force/torque readings) to recognize 20 grasped objects, using a CNN and $\sim\!1000$ training samples~\cite{Schmitz2014}. There have since been several other studies with taxel-based sensors~\cite{Cao2015,Meier,Baishya2016,Kwiatkowski2017}.

More recently, the Gelsight optical tactile sensor has found a natural match with deep learning, beginning with applications of CNNs to shape-independent hardness perception~\cite{Yuan2017c} and grasp stability~\cite{Calandra2017}, then considering visuo-tactile tasks for surface texture perception~\cite{Yuan2017b} and slip detection~\cite{Li2018}. The original Gelsight domed form-factor has been modified to a slimmer design for better integration with a two-digit gripper~\cite{Donlon2018}, enabling study of tactile grasp readjustment~\cite{Hogan2018} and visuo-tactile grasping~\cite{Calandra2018}. The majority of these studies use a CNN trained with several thousand examples, sometimes including an LSTM layer for sequential information. 

In this work, we consider tactile edge perception and contour following. In humans, edges and vertices are the most salient local features of 3D-shape~\cite{Plaiser2009,Lederman1997}, and are thus a key sensation for artificial tactile systems such as robot hands or prosthetics~\cite{PonceWong2013a}. Work in robotic tactile contour following dates back a quarter century~\cite{Berger1988,Berger1991,Chen1995}, with a more recent approach adopting a control framework for tactile servoing~\cite{Li2013}. However, these studies have relied on applying image processing techniques ({\em e.g.} image moments) to planar taxel arrays. For curved biomimetic sensors such as the iCub fingertip, another approach is to use a non-parametric probabilistic model of the taxel outputs~\cite{Martinez-Hernandez2013b,Martinez-Hernandez2013a,Martinez-Hernandez2017}.

The TacTip tactile sensor has been used for contour following, using a combination of servo control and a probabilistic model of the pin displacements~\cite{Lepora2017}. After tuning the control policy, the robot could tap around shapes such as a circle, volute and spiral. However, the trajectories~\cite[Figs 7-10]{Lepora2017} were not robust to parameter changes and failed when applied to sliding rather than tapping motion. Controlled sliding using touch is a challenge because the sensing surface undergoes motion-dependent shear due to friction against the object surface~\cite{Chen2018a}. Training data thus differs from the sensor output during the task, which will cause supervised learning methods to fail unless shear invariance is somehow applied. 


\section{METHODS}
\label{sec:3}

\subsection{Robotic system: Tactile sensor mounted on a robot arm}
\label{sec:3a}

\subsubsection{Tactile sensor}
\label{sec:3a1}

We use an optical biomimetic tactile sensor developed in Bristol Robotics Laboratory: the TacTip~\cite{Chorley2009,Ward-Cherrier2018}. The version used here is 3D-printed with a 40\,mm-diameter hemispherical sensing pad  (Fig.~\ref{fig:1}b) and 127 tactile pins in a triangular hexagonal lattice (Fig.~\ref{fig:1}c). Deformation of the sensing pad is imaged with an internal camera (ELP USB 1080p module; used at $640\times480$ pixels and $30$\,fps). The pin deflections can accurately characterize contact location, depth, object curvature/sharpness, edge angle, shear and slip. For more details, we refer to recent studies with this tactile sensor~\cite{Ward-Cherrier2018,Ward-Cherrier2017a,Ward-Cherrier2017b,Ward-Cherrier2016a,Cramphorn2018,Cramphorn2017,Cramphorn2016,Lepora2017,Lepora2016,Lepora2016a,Lepora2015,Pestell2018,Aquilina2018,James2018}.

\subsubsection{Robot arm mounting}
\label{sec:3a2}

The TacTip is mounted as an end-effector on a 6-DoF robot arm (IRB 120, ABB Robotics). The removable base of the TacTip is bolted onto a mounting plate attached to the rotating (wrist) section of the arm, then the other two modular components (central column and tip) are attached by bayonet fittings (Fig.~\ref{fig:1}a). 

\subsubsection{Software infrastructure}
\label{sec:3a3}

Our integrated sensorimotor framework has four components (Fig.~\ref{fig:2}): (1) Stacks of tactile images are collected in the OpenCV library, then preprocessed and either saved (for use in training) or used directly for prediction; 
(2)~These images are first cropped and subsampled to $(128\times128)$-pixel grey-scale images (Fig.~\ref{fig:1}c) and then passed to the Deep Learning system in Keras; (3) The resulting predictions are passed to control and visualization software in MATLAB; (4) The computed change in sensor pose is sent to a Python client that interfaces with a RAPID API for controlling the robot arm. Training is on a Titan Xp GPU hosted on a Windows 10 PC. {\color{black}The components run in real-time on the tasks (cycle time: $\sim\!1$ms for prediction and $\sim\!100$ms for control).}

\subsection{Deep learning system}
\label{sec:3b}

\begin{figure}[t!]
	\centering
	\includegraphics[width=0.95\columnwidth,trim={100 230 140 170},clip]{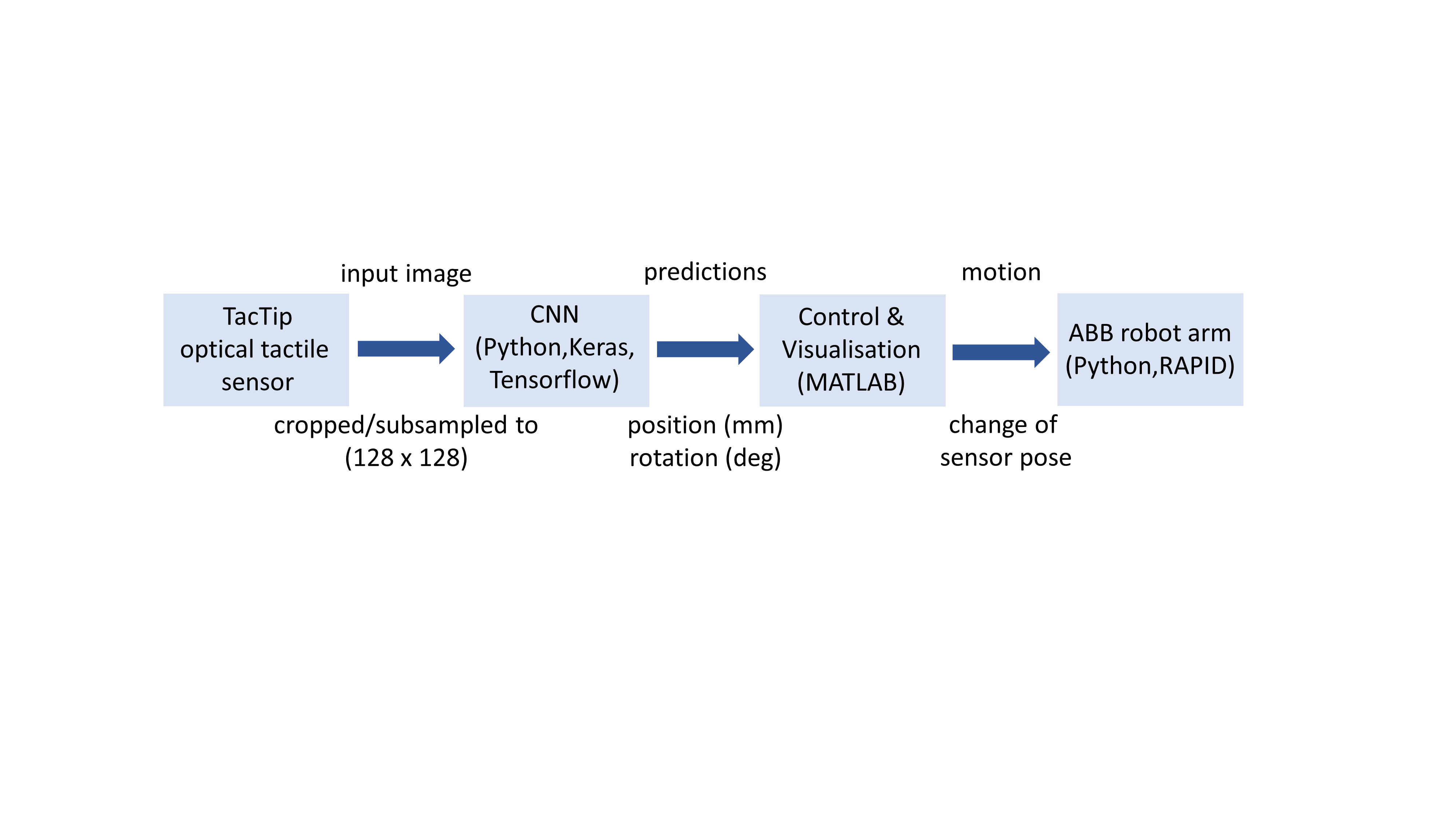}
	\caption{System and software infrastructure, from the TacTip optical tactile sensor to controlling the ABB robot arm.}
	\label{fig:2}
	\includegraphics[width=0.95\columnwidth,trim={200 50 180 10},clip]{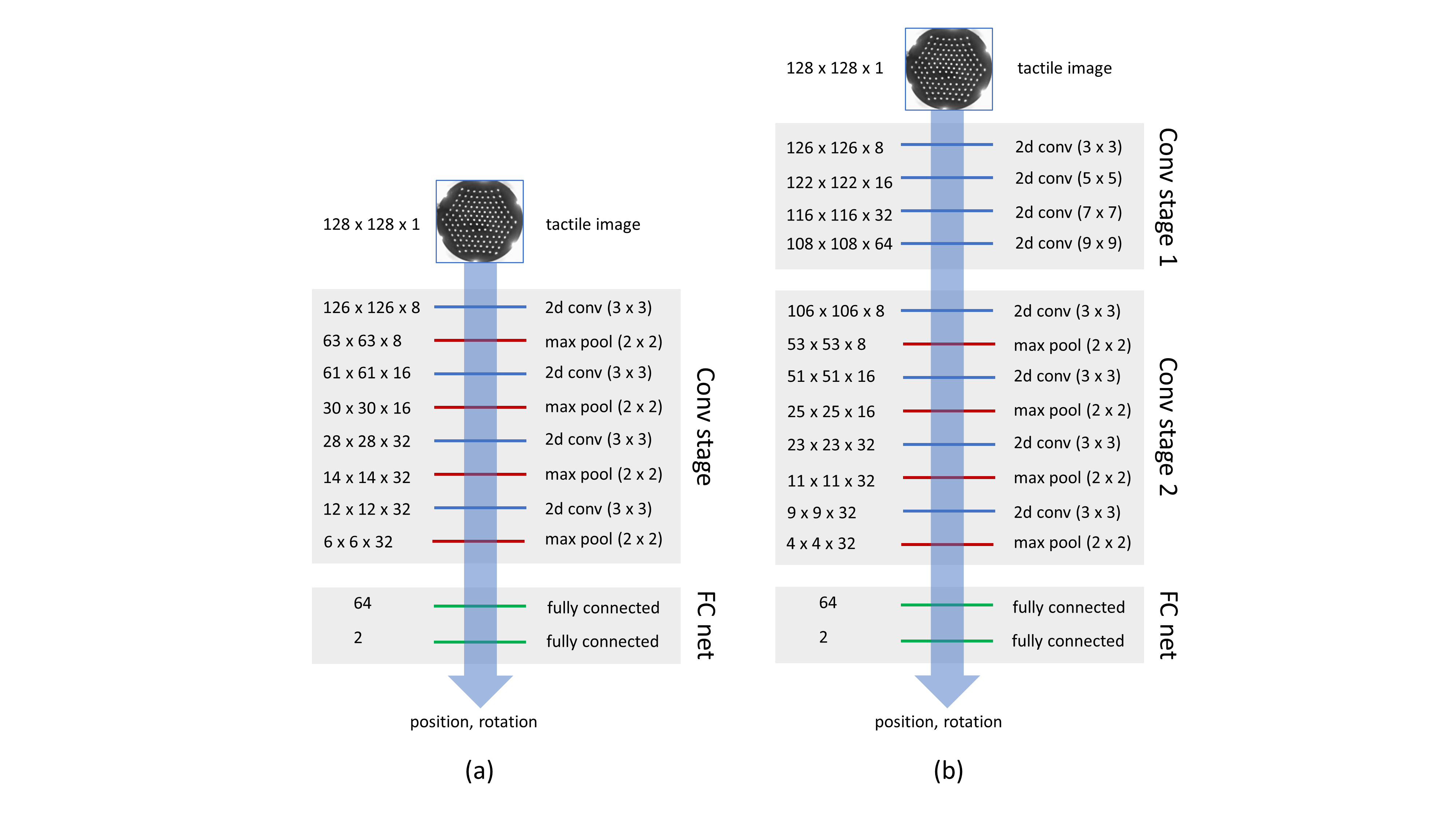}
	\caption{The two CNN architectures used in this work. The first (a) is based on a standard CNN pattern~\cite{Goodfellow2016} and the second (b) has an additional input stage without max pooling.}
	\label{fig:3}
	\vspace{-1em}
\end{figure}

Two types of CNN architecture are used here (Fig.~\ref{fig:3}): the first for tapping-based experiments; and the second, more complex architecture, to cope with the additional challenges associated with continuous-contact sliding motion.

The first architecture (Fig.~\ref{fig:3}a) passes a $128\times 128$ grey-scale image through a sequence of convolutional and max-pooling layers to generate a high-level set of tactile features. These features are then passed through a fully-connected regression network to make predictions. This configuration is based on a simple convolutional network pattern~\cite{Goodfellow2016}, which was scaled and regularized by restricting the number of filters/ReLUs in each layer and using a dropout of 0.25 between the convolutional stage and fully-connected net. 

The second architecture (Fig.~\ref{fig:3}b) was introduced to cope with the effects of non-linear shear on the sensor pin positions during continuous sliding motion. Under this type of motion, the pins tend to be displaced from the positions they occupy normally during tapping and thus their relative positions convey more useful information than their absolute positions. 

\begin{figure}[t!]
	\centering
	\includegraphics[width=0.48\columnwidth,trim={0 12 0 0},clip]{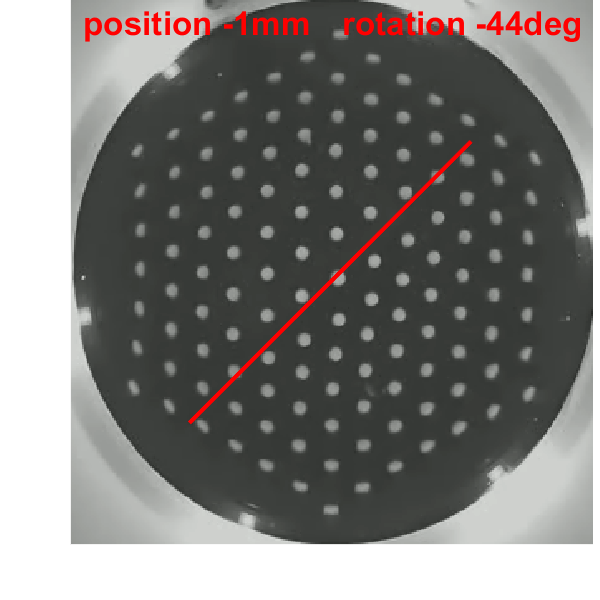}
	\includegraphics[width=0.48\columnwidth,trim={0 12 0 0},clip]{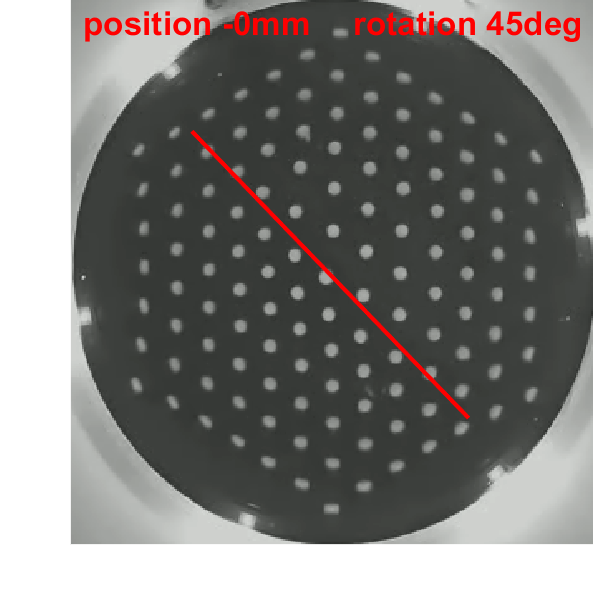}
	\caption{\color{black}Example tactile images from tapping on the disk. The angle is visible from the broader pin spacing above the edge. Further examples are given in the supplemental video.}
	\label{fig:4}
	\vspace{-1em}
\end{figure}

Initially, we tried to encourage the original network architecture (Fig.~\ref{fig:3}a) to make use of relative pin positions by using data augmentation to introduce randomly shifted copies of the sensor images into the training data (shifting each image randomly by 0-2\% in the horizontal and vertical directions on each presentation). However, this alone was not enough to achieve good performance for continuous-contact contour following around objects other than the disk.

We then extended the network by adding an initial convolution stage without any max-pooling layers before the original network (Fig.~\ref{fig:3}b), which did achieve better performance. Combined with the data augmentation, this allowed the network to learn broader features over larger groups of pixels, which allows the network to capture the spatial relationship between groups of adjacent pins. Once again, we used a dropout of 0.25 between the second convolutional stage and the fully-connected net to help over-regularize the model with respect to the validation and test data.

Both network architectures were trained using the Adam optimizer, with learning rate $10^{-4}$ and learning rate decay $10^{-6}$. {\color{black}All networks were trained from scratch, using the default Keras weight initializers (`glorot$\_$uniform').} Limiting the number of filters/ReLUs in each layer, using 0.25 dropout before the fully-connected net and early stopping (patience parameter 5) all helped prevent overfitting.


\subsection{Task: Tactile servoing along a contour}
\label{sec:3d}

Here we consider tasks in which a tactile sensor moves along a continuously-extended tactile feature such as the edge of an object, while rotating to maintain alignment with that feature. {\color{black}The control policy plans the next point to move along the contour around the object edge.}

These tasks are performed on a range of objects chosen for a variety of shapes and material properties: a $105\mm$-diameter circular disk, a tear-drop and a clover (all 3D-printed in ABS plastic); a lamina volute with radii of curvature $30, 40, 50, 60\mm$ in $90\deg$ segments and a $5\mm$-wide ridge in a volute spiral with radii of curvature $20, 30, 40, 50, 60\mm$ in $180\deg$ segments (both laser-cut from acrylic sheet); and two objects from the YCB object set~\cite{CAlli2015}, one chosen to be compliant (a soft rubber brick) and the other irregular (a plastic banana). Only the circular disk is used to gather training data (at the 12 o'clock position).

\begin{figure}[t!]
	\centering
	\begin{tabular}[b]{@{}cc@{}}
		{\bf (a) Position error} & {\bf (b) Rotation error}\\
		\includegraphics[width=0.45\columnwidth,trim={145 0 15 10},clip]{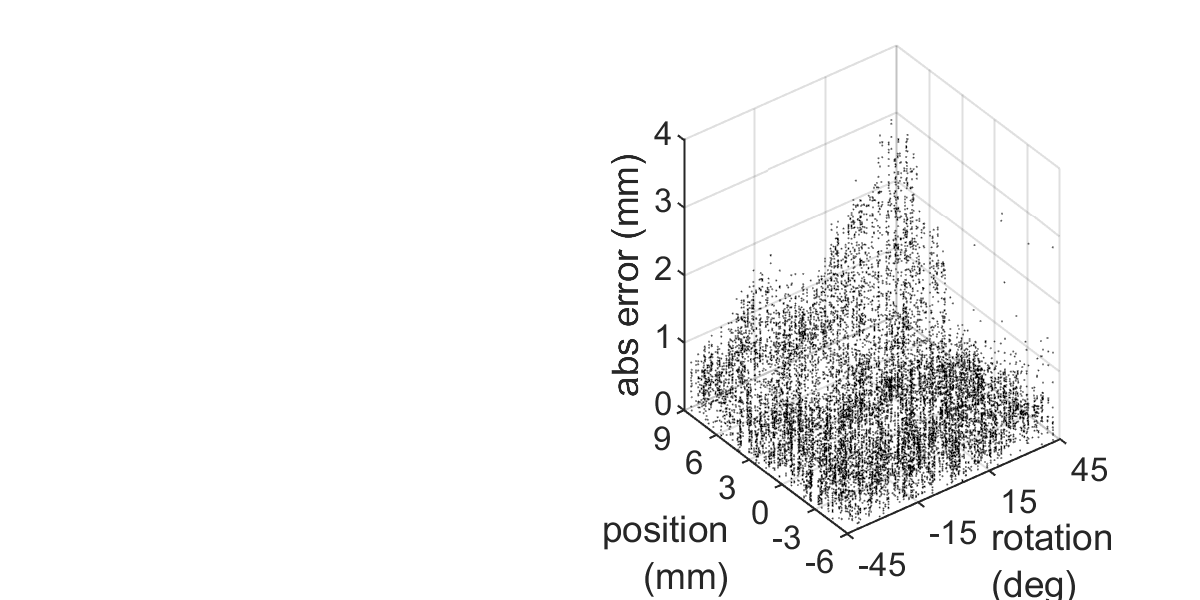} &
		\includegraphics[width=0.45\columnwidth,trim={15 0 145 10},clip]{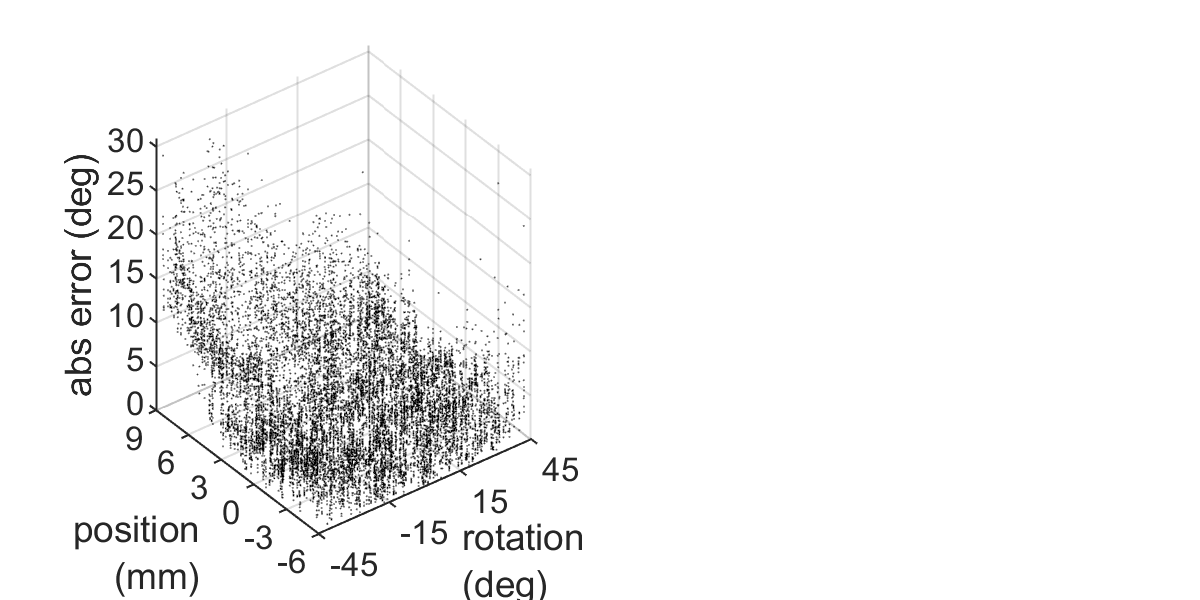} \\
	\end{tabular}
	\caption{Radial position and angular rotation prediction errors for the CNN in Fig.~\ref{fig:3}a, over 2000 random taps of 7 frames.}
	\vspace{1em}
	\label{fig:5}
	\begin{tabular}[b]{@{}cc@{}}
		{\bf (a) Position error} & {\bf (b) Rotation error}\\
		\includegraphics[width=0.45\columnwidth,trim={0 0 0 0},clip]{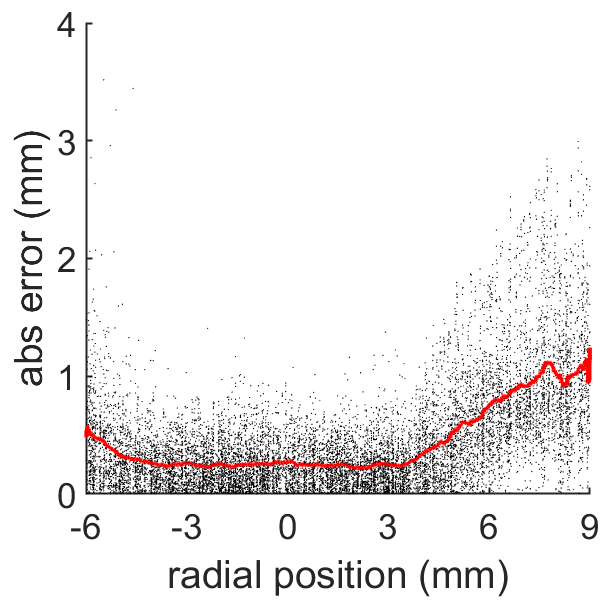} &
		\includegraphics[width=0.45\columnwidth,trim={0 0 0 0},clip]{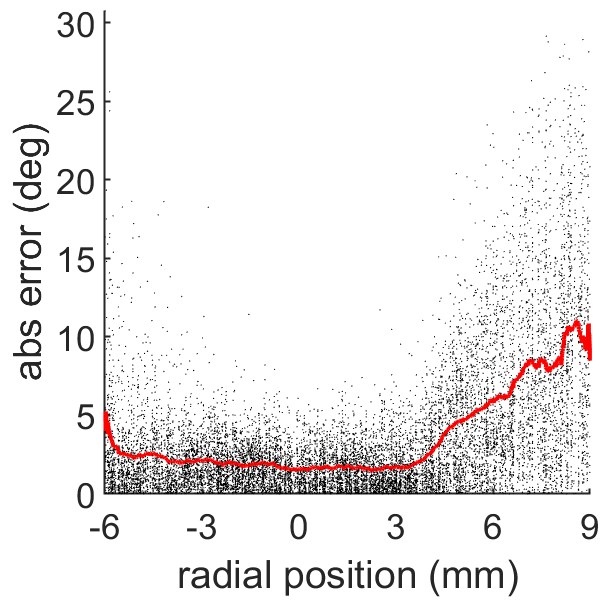} \\
	\end{tabular}
	\caption{CNN prediction errors from Fig.~\ref{fig:5} over radial position only. The red curve is a mean calculated from a 500-point (1/28 of the range) moving-average filter.}
	\label{fig:6}
	\vspace{-1em}
\end{figure}

{\color{black}We use a policy based on tactile servoing of the sensor pose to along the predicted edge orientation $(r,\theta)$, with three components~\cite{Lepora2017}: (i) a radial move $\Delta r$ along the (predicted) normal to the edge; (iii) an axial rotation of the sensor $\Delta\theta$; and (iii) a translation $\Delta e$ along the (predicted) tangent to the edge. This can be represented by a proportional controller
\begin{equation}
\label{eq:1}
\Delta r = g_r\left(r_0 - r\right),\hspace{1em} \Delta\theta = g_{\theta} \left(\theta_0-\theta\right),
\end{equation}
with unit gains $(g_r,g_\theta)=(1,1)$, set-point $(r_0,\theta_0)=(0\mm, 0\deg)$ (the edge pose in these coordinates) and we choose a default step $\Delta e=3$\,mm used before in ref.~\cite{Lepora2017}. 

This control policy plans points for following an edge or contour. In the simple case considered here, the actions $(\Delta r,\Delta\theta)=(-r,-\theta)$ correspond to the training-data labels and the tangential step $\Delta e$ is assumed sufficiently small to not lose a curved edge. The policy may then be considered as end-to-end from the tactile images to controlled actions.}

\subsection{Data collection}
\label{sec:3c}

For training, the tactile robotic system samples a local region of the edge of one object ($105\mm$ disk at 12 o'clock position) over a range of radial positions $r$ and axial (roll) rotation angles $\theta$. Here we used 2000 uniform random tapping contacts sampled over ranges $(-6,+9)\mm$ and $(-45,+45)\deg$, with each tap from $\sim1.5\mm$ above the object, down $5\mm$, over $\sim0.7\sec$ ($\sim20$ frames). The origin $(0\mm, 0\deg)$ has the sensor tip centred on the edge with camera and pin lattice aligned normal to the edge. 

The training set was split into 1600 samples for learning the weights and 400 for hyperparameter optimization. Another dataset of 2000 contacts over new random positions and angles was used for evaluating perceptual performance. {\color{black}We used the 7 frames around the peak displaced frame (measured by the change in RMS pixel intensity from the initial frame) to capture data near the deepest contact part of the tap, including some variation over depth but excluding non-contact data.} The images were then cropped and subsampled to a $(128\times128)$-pixel region containing the pins (Fig.~\ref{fig:4}).

There is modest scope for improving the results by fine-tuning these experiment parameters.  However, trial and error (and experience with the experiment~\cite{Lepora2017}) showed these to be reasonable and natural choices. The only non-obvious choice was to include a random shift between $\pm 1\deg$ of the sensor yaw/pitch in all data: this reduced specialisation in the trained network to small but noticeable non-normal sensor alignments that otherwise biased the angle predictions.



\section{RESULTS}
\label{sec:4}

\subsection{End-to-end edge perception from tactile images}
\label{sec:4a}

In a major departure from recent work with the TacTip optical biomimetic sensor, here we predict the percepts directly from tactile images with a deep CNN. Prior work with this sensor has used specialised preprocessing to detect then track the pin positions~\cite{Ward-Cherrier2018,Ward-Cherrier2017a,Ward-Cherrier2017b,Ward-Cherrier2016a,Cramphorn2018,Cramphorn2017,Cramphorn2016,Lepora2017,Lepora2016,Lepora2016a,Lepora2015,Pestell2018,Aquilina2018,James2018}. Here, this preprocessing is subsumed into the trained neural network. 

We report the performance of the first CNN architecture (Fig.~\ref{fig:3}a). During our preliminary investigations, we found that networks with more filters/ReLUs in each layer and less regularization achieved better performance on the validation and test data collected at the same point on the disk as where the training data was collected. However, they failed to generalize well to other regions of the disk or to other objects. Over-regularizing the network beyond the point required for good generalization on the test data helped solve this problem and produced models that perform well on a broader range of tapping-based contour following tasks.

The overall perceptual performance is then most accurate near the central positions for all rotations (Fig.~\ref{fig:5}). In this region ($-3$ to $+3\mm$), errors are generally less than $1\mm$ and $5\deg$. Overall, the contacts are less informative further from the edge ($9\mm$ into free space; $-6\mm$ onto the disk), consistent with the edge being no longer visible in those tactile images. Considering only the dependence on position (Fig.~\ref{fig:6}), the mean absolute errors are $\sim\!0.3\mm$ and $\sim\!2\deg$ in the central region where perception is most accurate (red curve), appropriate for the contour-following tasks below.

\begin{figure}[h!]
	\centering
	\begin{tabular}[b]{@{}cc@{}}
		{\bf (a) Taps: Initial contact} & {\bf (b) Taps: Step size} \\
		\includegraphics[width=0.45\columnwidth,trim={25 15 5 5},clip]{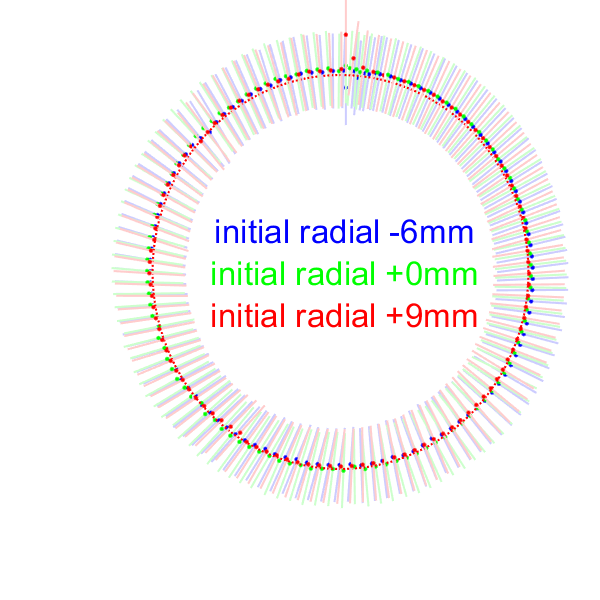} &
		\includegraphics[width=0.45\columnwidth,trim={25 15 5 5},clip]{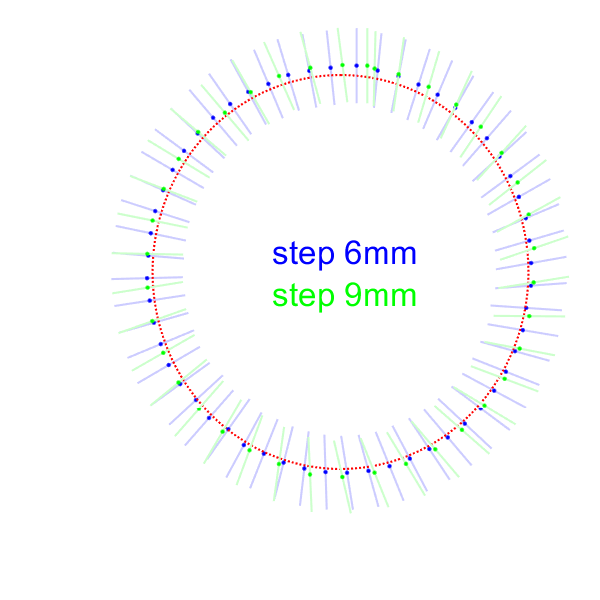} \\
		{\bf (c) Taps: Contact radius} & {\bf (d) Taps: Contact depth} \\
		\includegraphics[width=0.45\columnwidth,trim={25 15 5 0},clip]{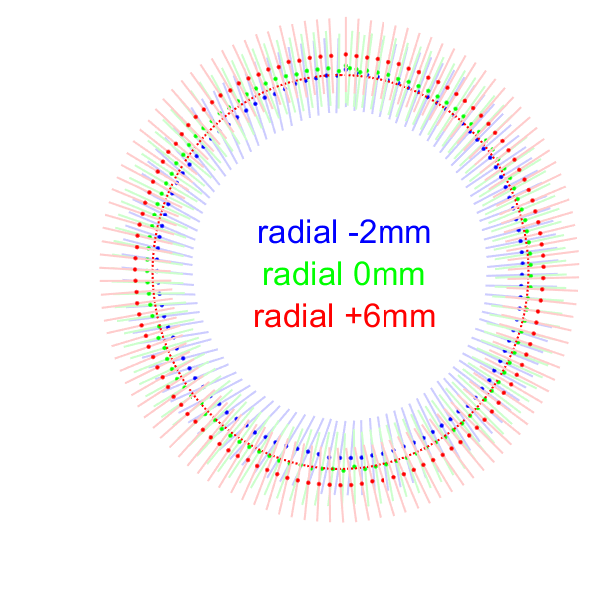} &
		\includegraphics[width=0.45\columnwidth,trim={25 15 5 0},clip]{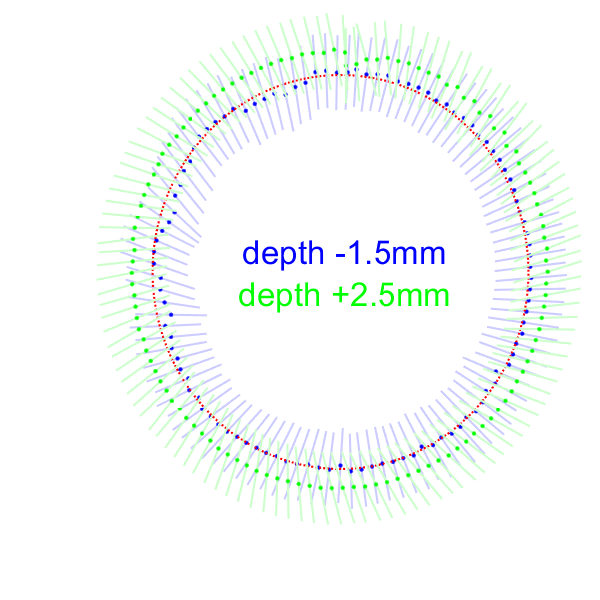} \\
	\end{tabular}
	\caption{Robust tapping around a uniform disk to the contour changes in Table~\ref{tab:1} (trained at the 12 o'clock position).}
	\label{fig:7}
	\vspace{1em}
	\centering
	\begin{tabular}[b]{@{}cc@{}}
		{\bf (a) Taps: Volute lamina} & {\bf (b) Taps: Spiral ridge} \\
		\includegraphics[width=0.45\columnwidth,trim={10 15 0 5},clip]{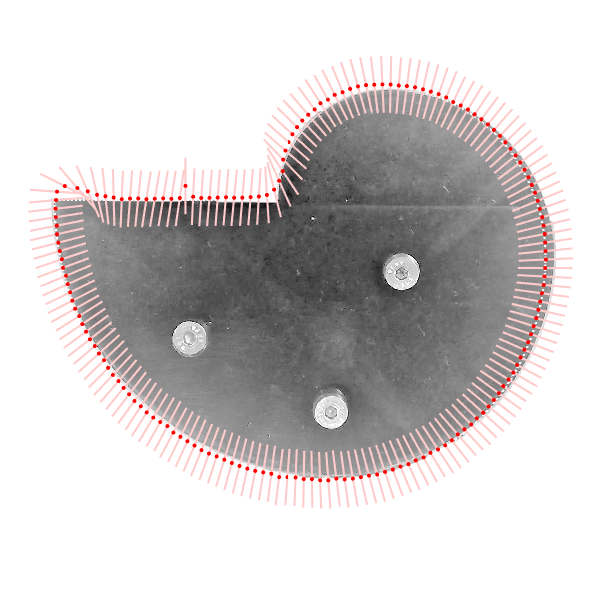} &
		\includegraphics[width=0.45\columnwidth,trim={10 15 0 5},clip]{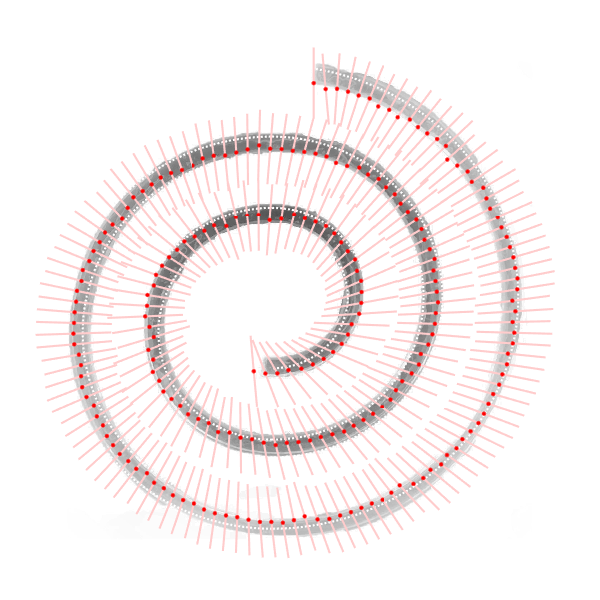} \\
		{\bf (c) Taps: Foil} & {\bf (d) Taps: Clover} \\
		\includegraphics[width=0.45\columnwidth,trim={10 20 0 15},clip]{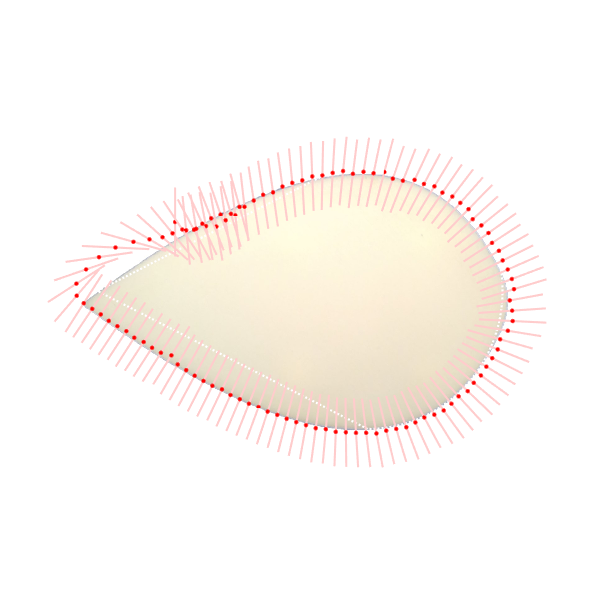} &
		\includegraphics[width=0.45\columnwidth,trim={10 20 0 15},clip]{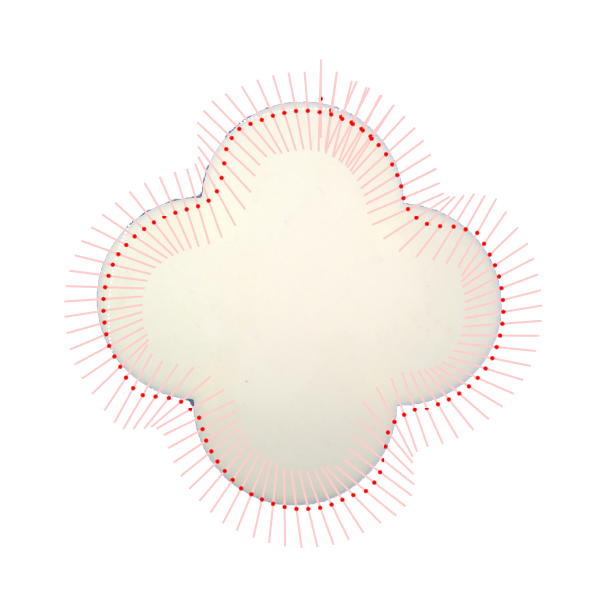} \\
		{\bf (e) Taps: Compliant object} & {\bf (f) Taps: Irregular object} \\
		\includegraphics[width=0.4\columnwidth,trim={20 30 20 40},clip]{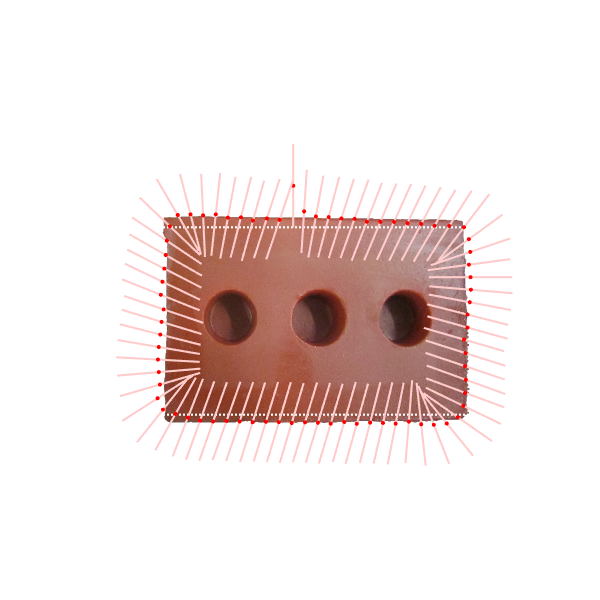} &
		\includegraphics[width=0.45\columnwidth,trim={15 40 15 45},clip]{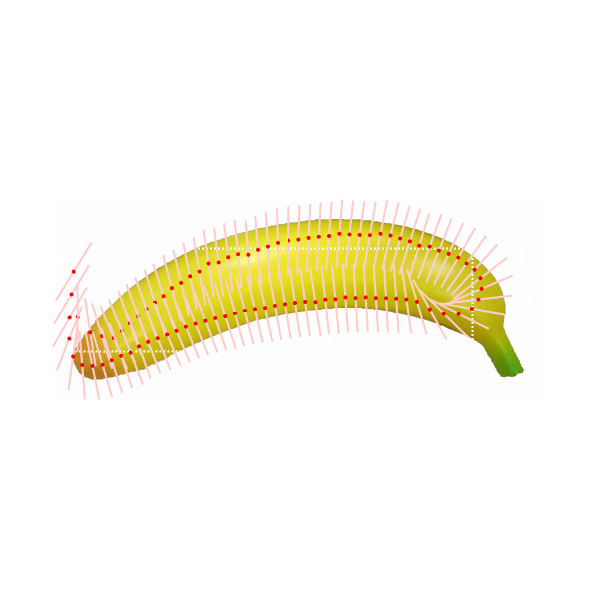} \\
	\end{tabular}
	\vspace{0em}
	\caption{Robust tapping around non-uniform objects (trained with taps at the 12 o'clock position on the disk).}
	\label{fig:8}
\end{figure}

\begin{figure}[h!]
	\centering
	\begin{tabular}[b]{@{}cc@{}}
		{\bf (a) Slide: Initial contact} & {\bf (b) Slide: Step size} \\
		\includegraphics[width=0.45\columnwidth,trim={25 15 5 5},clip]{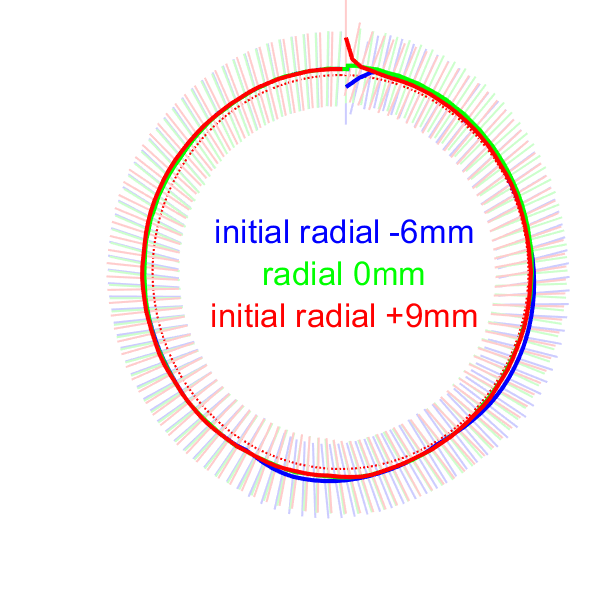} &
		\includegraphics[width=0.45\columnwidth,trim={25 15 5 5},clip]{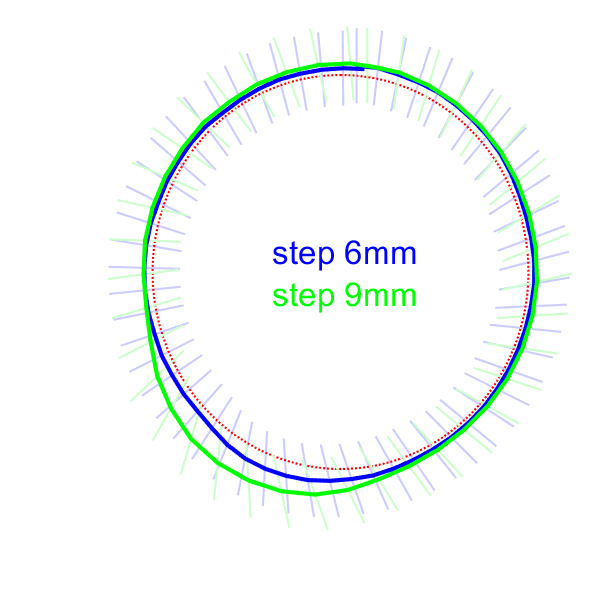} \\
		{\bf (c) Slide: Contact radius} & {\bf (d) Slide: Contact depth} \\
		\includegraphics[width=0.45\columnwidth,trim={25 15 5 0},clip]{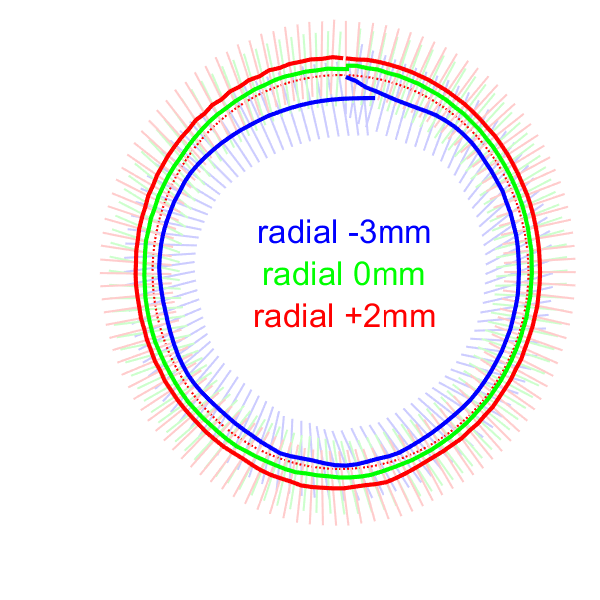} &
		\includegraphics[width=0.45\columnwidth,trim={25 15 5 0},clip]{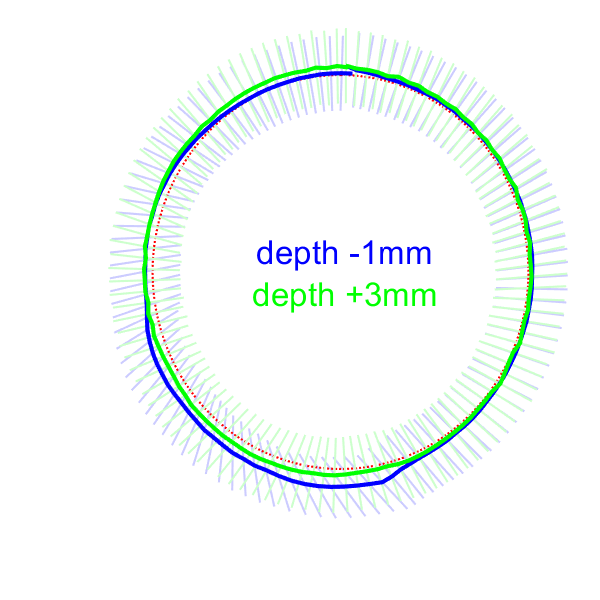} \\
	\end{tabular}
	\caption{Robust sliding around a disk to the contour changes in Table~\ref{tab:1} (trained with taps at the 12 o'clock position).}
	\label{fig:10}
	\vspace{1em}
	\centering
	\begin{tabular}[b]{@{}cc@{}}
		{\bf (a) Slide: Volute} & {\bf (b) Slide: Spiral ridge} \\
		\includegraphics[width=0.45\columnwidth,trim={10 15 0 5},clip]{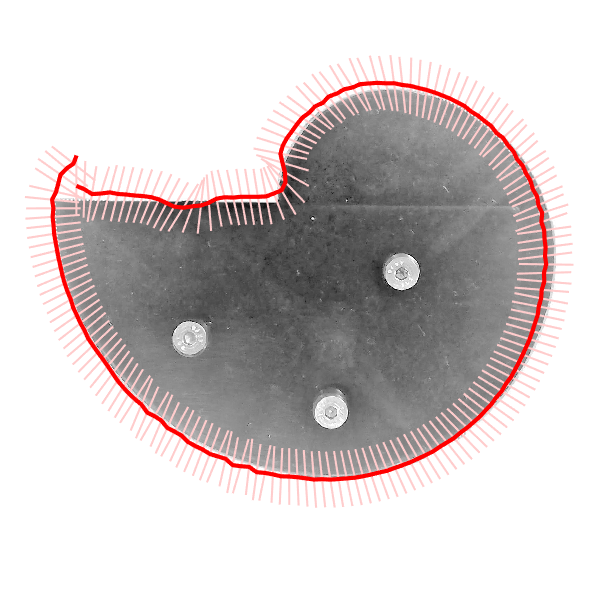} &
		\includegraphics[width=0.45\columnwidth,trim={10 15 0 5},clip]{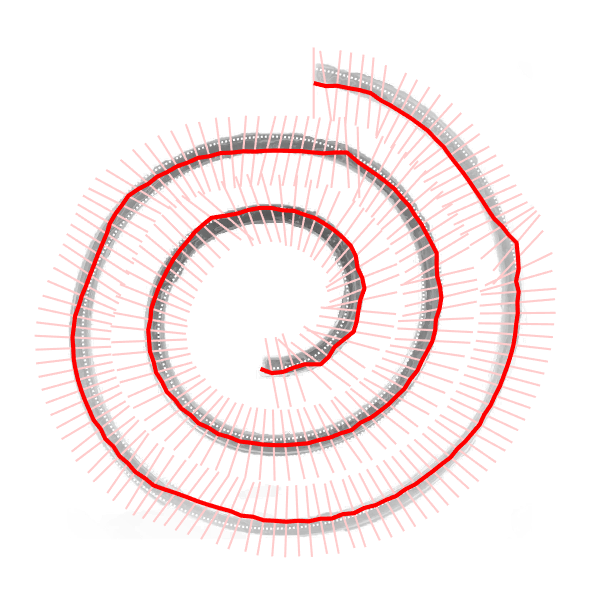} \\
		{\bf (c) Slide: Foil} & {\bf (d) Slide: Clover} \\
		\includegraphics[width=0.45\columnwidth,trim={10 20 0 15},clip]{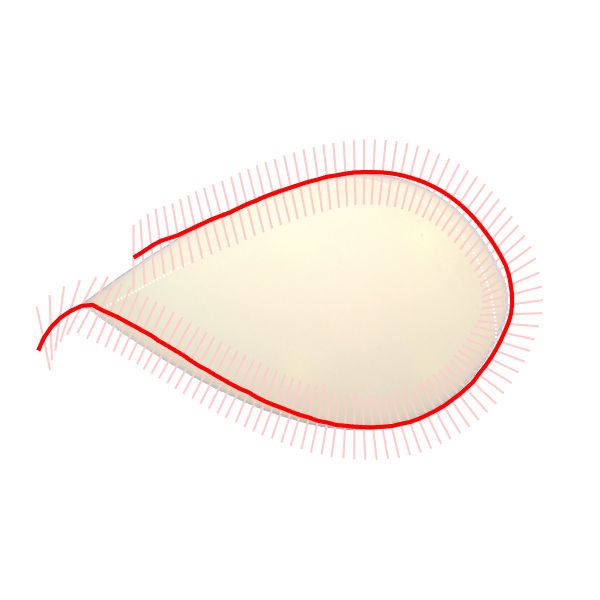} &
		\includegraphics[width=0.45\columnwidth,trim={10 20 0 15},clip]{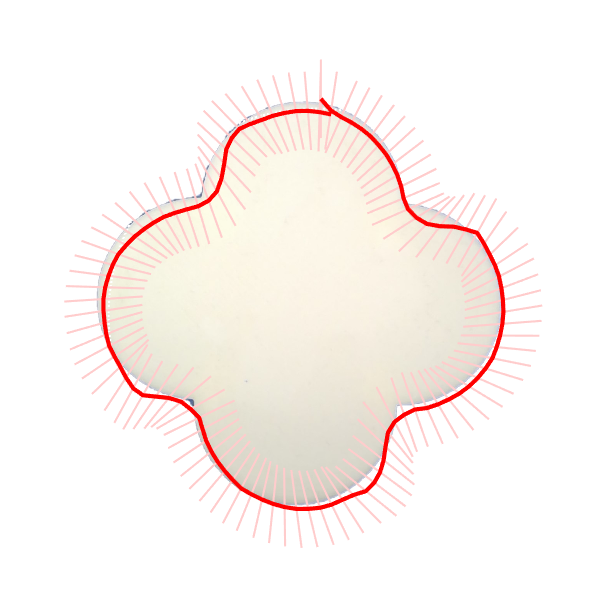} \\
		{\bf (e) Slide: Compliant object} & {\bf (f) Slide: Irregular object} \\
		\includegraphics[width=0.4\columnwidth,trim={20 30 20 40},clip]{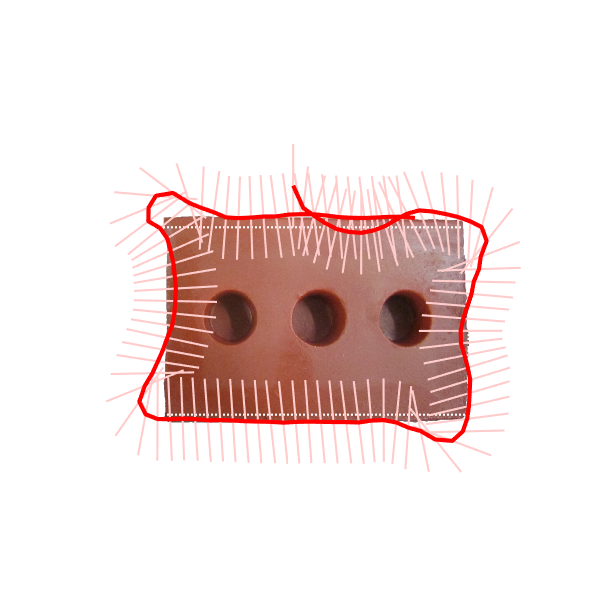} &
		\includegraphics[width=0.45\columnwidth,trim={15 40 15 45},clip]{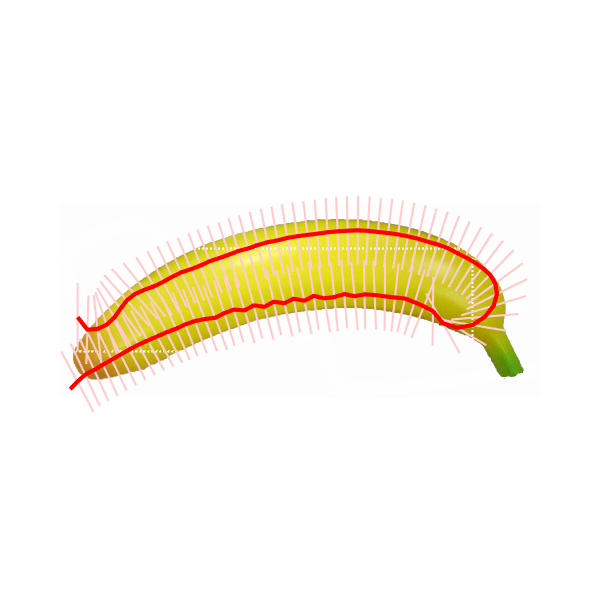} \\
	\end{tabular}
	\vspace{0em}
	\caption{Robust sliding around non-uniform objects (trained with taps at the 12 o'clock position on the disk).}
	\label{fig:11}
\end{figure}

\subsection{Robust contour following around a disk}
\label{sec:4b}

The deep CNN model for predicting edge pose angle and radial position is now applied to contour following around the disk with tapping contacts. A servo control policy (Eq.~\ref{eq:1}) plans contact points to maintain the relative pose to the edge.

The completed trajectories are near-perfect circles around the disk under a range of conditions (Fig.~\ref{fig:7}; Table~\ref{tab:1}). {\color{black}Trajectories are repeatable, as indicated by running the experiment from different starting positions relative to the edge (Fig.~\ref{fig:7}a), with a small offset in the radial displacement (0-1\,$\mm$) and sensor angle (6-9\,$\deg$) relative to the edge.} Increasing the policy step size from $3\mm$ keeps the trajectory within $2\mm$ of the edge, but offsets the angle by another 4-11\,$\deg$ consistent with the sensor moving from its predicted location (Fig.~\ref{fig:7}b; $\Delta e\!=\!6,9\mm$). Taking a set-point radius inside or outside the edge keeps a circular trajectory without contributing to the angle offset (Fig.~\ref{fig:7}c; $r_0\!=\!-3,+6\mm$).

The circular trajectories are also robust to changing the tapping depth (Fig.~\ref{fig:7}d, depth change $\Delta=-1.5,+2.5\mm$), from shallow taps ($2.5\mm$ above, down $5\mm$) to deep taps ($-1.5\mm$ above, down $5\mm$). Shallow taps advanced the sensor angle by $8\deg$ and deep taps lag the angle by $20\deg$. The additional is because the task data is at a different depth from that used to train the model; although the model is most accurate at the training depth, its performance has declined gracefully to still complete the contour following task.

These results are a major improvement over those obtained in a previous study on the same task with a probabilistic model of pin displacements~\cite{Lepora2017} (comparison in Table~\ref{tab:1}), which failed to complete the task in many circumstances. 

\begin{table}[!b]
	\vspace{-1em}
	\renewcommand{\arraystretch}{1}
	\centering
	\setlength{\tabcolsep}{0.33em}
	\begin{tabular}{cc|cc}
		{\bf experiment} &\renewcommand{\arraystretch}{1}\begin{tabular}{c} {\bf parameter} \\ {\bf variation} \end{tabular} &
		\renewcommand{\arraystretch}{1}\begin{tabular}{c} {\bf probabilistic} \\ {\bf model~\cite{Lepora2017}} \end{tabular} &
		{\bf deep CNN} \\
		\hline
		tapping contact & $r_{\rm init}\!=\!-6\mm$  & $1\mm$, $23\deg$                & $0\mm$, $6\deg$ \\
		(Fig.~\ref{fig:7}a) & $r_{\rm init}\!=\!0\mm$  & $2\mm$, $25\deg$    & $1\mm$, $6\deg$ \\
	    & $r_{\rm init}\!=\!9\mm$ & $2\mm$, $41\deg$    & $0\mm$, $9\deg$ \\
		\hline
		tapping contact & $\Delta e\!=\!6$$\mm$ & $2\mm$, $25\deg$  & $1\mm$, $10\deg$ \\
		(Fig.~\ref{fig:7}b) & $\Delta e\!=\!9$$\mm$ & fail & $1\mm$, $17\deg$ \\ 
		\hline
		tapping contact & $r_0\!=\!-2$$\mm$ & fail & $1\mm$, $6\deg$ \\
		(Fig.~\ref{fig:7}c) & $r_0\!=\!+6$$\mm$ & $3\mm$, $16\deg$             & $3\mm$, $7\deg$ \\
		\hline
		tapping contact & $\Delta\!=\!-1.5$$\mm$  & fail & $0\mm$, $8\deg$ \\
		(Fig.~\ref{fig:7}d) & $\Delta\!=\!+2.5$$\mm$  & $4\mm$, $25\deg$          & $3\mm$, $20\deg$ \\
		\hline
		sliding contact & $r_{\rm init}\!=\!-6\mm$  & fail & $1\mm$, $15\deg$ \\
		(Fig.~\ref{fig:10}a) & $r_{\rm init}\!=\!0\mm$  & fail & $1\mm$, $12\deg$ \\
		& $r_{\rm init}\!=\!9\mm$ & fail    & $1\mm$,$11\deg$ \\
		\hline
		sliding contact & $\Delta e\!=\!6$$\mm$ & fail		     & $1\mm$, $15\deg$ \\
		(Fig.~\ref{fig:10}b) & $\Delta e\!=\!9$$\mm$ & fail             & $3\mm$, $34\deg$ \\
				\hline
		sliding contact & $r_0\!=\!-3$$\mm$ & fail & $2\mm$, $25\deg$ \\
		(Fig.~\ref{fig:7}c) & $r_0\!=\!+2$$\mm$ & fail & $2\mm$, $9\deg$ \\
		\hline
		sliding contact & $\Delta\!=\!-1$$\mm$  & fail           & $1\mm$, $18\deg$ \\
		(Fig.~\ref{fig:7}d) & $\Delta\!=\!+3$$\mm$  & fail          & $1\mm$, $4\deg$ \\ 		
	\end{tabular}
	\caption{Accuracy of exploration around the disk, showing the mean absolute errors of radial position and rotation angle for the trajectories in Figs~\ref{fig:7},\ref{fig:10}. A comparison is shown with the original probabilistic model~\cite{Lepora2017} for each experiment.}
	\label{tab:1}
\end{table}

\subsection{Robustness to non-uniform object shapes}

To demonstrate further robustness, the task is extended to four fabricated planar shapes chosen to have non-uniform curvature (a volute lamina, spiral ridge, tear drop and clover) and two household objects from the YCB object set~\cite{CAlli2015}: one compliant (a soft rubber brick) and one irregular (a plastic banana). Task completion shows generalization to novel contours differing from the disk edge used in training.

For the four fabricated shapes, the completed trajectories match the object shapes (Figs~\ref{fig:8}a-d). When the radius of curvature is close to that of the disk, the sensor angle aligns to the edge normal. For smaller or larger curvatures, there is offset in the sensor angle (advancing for smaller and lagging for higher). The large changes in orientation at corners on the volute and foil cause overshoots, but the task still completes.

 
The task also completes on the compliant object~(Fig.~\ref{fig:8}e, rubber brick). Again, the policy advances the edge angle on the straight sections. There is less overshoot at the compliant corners than those of the rigid objects; however, the angle offset on the straight edges would make turning easier.

The task does not quite complete on the irregular object~(Fig.~\ref{fig:8}f, banana). The task was challenging because there are only slight ridges and the height varies by a few millimetres. However, the policy managed to traverse most of the object, failing only at the tip where there is both a sharp change in orientation and no well-defined edge to follow.

\subsection{Robustness to sliding contact}

A far more demanding test of robustness is to use a continuous sliding motion with the same training data from tapping the disk. We repeat the above experiments, dropping the sensor $3\mm$ ($1.5\mm$ into the object) and collecting data between the exploratory movements ($\sim\!0.15\sec$ duration; $\sim\!5$ frames). The second CNN architecture (Fig.~\ref{fig:3}b) was used, which has an initial convolution stage without any max-pooling layers to help generalize over tactile features. 

Under a sliding motion, the task completion was robust to changes in starting point~(Fig.~\ref{fig:10}a), step size (Fig.~\ref{fig:10}b; $\Delta e\!=\!6,9\mm$), set-point radius (Fig.~\ref{fig:10}c; $r_{\rm fix}=-3,+2\mm$) and contact depth (Fig.~\ref{fig:10}d; $-1\mm$ shallower, $+3\mm$ deeper). The range of depths where the task completes is likely greater, but concerns about damaging the sensor limited further testing. The angle offset improved to $4\deg$ with the deepest contact, which is the best overall trajectory. 

Three objects (the spiral, clover and compliant brick) were successfully traversed with sliding (Figs~\ref{fig:11}b,d,e). The two rigid objects have similar sliding and tapping trajectories~({\em c.f.}~Figs~\ref{fig:8}b,d and \ref{fig:11}b,d). For the rubber brick, the angle offset on the straight edges appears less than for tapping motion, but there is a larger overshoot around the corners  ({\em c.f.}~Figs~\ref{fig:8}e and \ref{fig:11}e). We interpret this as a consequence of the sliding motion inducing a shear of the sensing surface that makes corners more challenging.

{\color{black}The policy failed on the other three objects (the volute, tear drop and banana) at the sharp corners (Figs~\ref{fig:11}a,c,f), but was successful otherwise. The first two were objects with large overshoots at the corners when tapping (Figs~\ref{fig:8}a,c), which appear to have caused the failure under the more demanding condition of sliding. The policy also failed at the tip of banana for both tapping (Fig.~\ref{fig:8}f) and sliding motion (Fig.~\ref{fig:11}f) where there is no well-defined edge to follow.}

\section{Discussion}

This work is the first application of deep learning to an optical biomimetic tactile sensor and the first to edge perception and contour following. We found robust generalization of the contour-following policy to tasks beyond which the model was trained, such as continuous sliding around compliant or irregular objects after training with taps on part of a disk.


We used two techniques to encourage generalization. The first was to anticipate nuisance variables to marginalise out and then either modify the data collection (e.g. training over shifts in yaw/pitch) or augment the data set with artificially generated data (e.g. randomly shifting frames). The second technique was to over-regularize the architecture to avoid specialization to the training task; this may be because it encourages the development of simpler features throughout the network. In both cases, we introduce inductive bias into the model to improve performance in situations different from its training. This is necessary because generalizing beyond the task a model is trained on cannot be reliably achieved by trying to validate on data from the original task

We emphasise the generalization from discrete tapping contacts to continuous sliding motion is a challenging test for the policy. Sliding causes a friction-dependent shear of the sensing surface that depends on the motion direction and recent history of the interaction~\cite{Chen2018a}. Hence, the tactile data during a task can differ in complex ways from those during training. Although this caused our previous statistical model to fail~\cite{Lepora2017}, the deep learning model performed robustly.

{\color{black}The principal failure mode of the deep learning model was on sharp corners under sliding motion. This is unsurprising, as the model was only trained on edge data from the disk, so corners are both outside its experience and give a singularity in the prediction. The model degraded gracefully, with corners successfully followed with a tapping motion, albeit with some overshoot, and also for sliding around reflex angles and the compliant object. In principle, this limitation could be solved by crafting a more complete policy that can predict points around corners, {\em e.g.} by training on corners of various angles. A complementary method would be to adapt the step size of the policy based on the predicted curvature of the object.}

In our view, the greatest benefit of using a deep CNN to learn a tactile control policy is its capability to generalize beyond the training data. Previous studies with the same biomimetic tactile sensor found good performance when the task and training were similar~\cite{Lepora2015,Lepora2016,Lepora2016a,Lepora2017}; however, we were aware that these results were fragile to small changes in the task ({\em e.g.} sensor orientation). Since practical applications of tactile sensing require robust performance in situations beyond those previously experienced, we expect this is a generic problem in robot touch that deep learning will solve.




{\em Acknowledgements:} We thank NVIDIA Corporation for the donation of the Titan Xp GPU used for this research



\newpage
\bibliographystyle{unsrt}
\bibliography{library}

\end{document}